\documentclass[10pt,twocolumn,letterpaper]{article}

\usepackage{ijcb}
\usepackage{times}
\usepackage{epsfig}
\usepackage{graphicx}
\usepackage{amsmath}
\usepackage{amssymb}
\usepackage[norule,symbol,perpage]{footmisc}
\usepackage{multirow}

\ijcbfinalcopy 

\ifijcbfinal\fi

\makeatletter
\def\ps@IEEEtitlepagestyle{
\def\@oddfoot{\mycopyrightnotice}
\def\@evenfoot{}
}
\def\mycopyrightnotice{
{ \hspace{-7mm}$^\dagger$ corresponding author}\\
{ 978-1-6654-3780-6/21/\$31.00 \copyright 2021 IEEE}
}

\makeatother

\begin{document}

\title{Structure Destruction and Content Combination for Face Anti-Spoofing}

\author{
 Ke-Yue Zhang,
 Taiping Yao,  
 Jian Zhang,
 Shice Liu,
 Bangjie Yin,
 Shouhong Ding$^\dagger$,
 Jilin Li \\
Youtu Lab, Tencent, Shanghai, China \\
{\tt\small \{zkyezhang,taipingyao,timmmyzhang,shiceliu,bangjieyin,ericshding,jerolinli\}@tencent.com}
}

\maketitle
\thispagestyle{empty}

\begin{abstract}
In pursuit of consolidating the face verification systems, prior face anti-spoofing studies excavate the hidden cues in original images to discriminate real person and diverse attack types with the assistance of auxiliary supervision. 
However, limited by the following two inherent disturbances in their training process: 1) Complete facial structure in a single image. 2) Implicit subdomains in the whole dataset, these methods are prone to stick on memorization of the entire training dataset and show sensitivity to non-homologous domain distribution. 
In this paper, we propose Structure Destruction Module and Content Combination Module to address these two limitations separately. 
The former mechanism destroys images into patches to construct a non-structural input, while the latter mechanism recombines patches from different subdomains or classes into a mixup construct. 
Based on this splitting-and-splicing operation, Local Relation Modeling Module is further proposed to model the second-order relationship between patches. We evaluate our method on extensive public datasets and promising experimental results to demonstrate the reliability of our method against the state-of-the-art competitors.
\end{abstract}

\let\thefootnote\relax\footnotetext{\mycopyrightnotice}

\vspace{-5mm}
\section{Introduction}
Face recognition techniques bring much convenience, but it also faces the safety issue since the face Presentation Attacks (PA) appear. 
Therefore, how to protect these systems from the presentation attacks promotes the techniques of face anti-spoofing (FAS).
In recent years, researchers put forward several hand-crafted features based and deep learning based methods for presentation attack detection. 
The latter is superior to the former since the Convolutional Neural Networks (CNN) hold strong representation abilities to tackle this issue. 
Firstly, some methods solved this problem by binary cross-entropy supervision~\cite{DeepBinary00}, which is easily overfitting on some uncritical details. 
Then, several methods utilized some auxiliary information to promote the performance of the model~\cite{additionInfo01,additionInfo02,additionInfo06}, such as facial depth map, remote-photoplethysmography signal, temporal information, \textsl{etc}. However, there are still two limitations.

1) \textit{Strong relevance to facial structure.} 
Face anti-spoofing is inherently concentrated on detail exploration, which is independent of global facial structure. 
Former methods inevitably introduce structural information since the whole face serves as the input. 
However, such structural information is not the essential features for FAS.

2) \textit{Implicit subdomains in the dataset.} It is the consensus that face anti-spoofing classifier should be robust to varying domains. 
However, the networks trained on the whole dataset may not utilize the implicit subdomain samples fully to boost the robustness of a classifier.

To address the above issues, we propose a novel face anti-spoofing framework named Destruction And Combination Network (DCN). The framework contains two mechanisms called Structure Destruction Module and Content Combination Module to solve above two limitations separately. Structure Destruction Module destroys the global structure of images then instructs the networks to concentrate on local details. Specifically, we construct non-structural input via destructing the face into several patches and permuting them in an arbitrary order. We clarify that local cues to distinguish real and fake are widely spread in the original images, making it feasible to predict each patch. Once detached, Content Combination Module further exploits the implicit subdomains in the dataset. We splice patches from different domains or classes into a mixed image, while the network should maintain correct class prediction for each part. 
Thus, our architecture excludes domain influence from the liveness judgment of the mixed image.

Based on the above recombination, we further propose Local Relation Modeling Module to explicitly model the second-order relationship between different patches. Concretely, we utilize cosine measurement to constrain the similarities between patches. To make consistent predictions between patches of the same class, this mechanism not only forces the network to extract local crucial patterns in a random context but also leads the network to exclude irrelevant domain properties, promoting generalization capability.

The main contributions are summarized as follows:

$\bullet$ A novel Destruction and Combination Network (DCN) is proposed for face anti-spoofing. By simultaneously destroying the facial structure in the original image and excluding domain knowledge in the dataset, our framework is reliable for extracting discriminative features thoroughly.

$\bullet$ Among patch-based frameworks, we further model the second-order relationship between patch pairs, promoting the capture of crucial information in each local region.

$\bullet$ We evaluate our method on extensive benchmarks. With superior performance and convincing visualization, we demonstrate the effectiveness of data recombination.

\section{Related Work}
\textbf{Face Anti-Spoofing Methods.}
Researchers have made great progress in the face anti-spoofing area recently. 
The development can be roughly divided into two stages. Early researches mainly utilized handcrafted feature descriptors, such as LBP~\cite{LBP00,LBP01,LBP02,LBP03}, HOG~\cite{HoG00,HoG01} and SIFT~\cite{SIFT00} and trained a traditional classifier for judgement. 
\cite{LBP00} leveraged information from different domains, such as HSV, to improve the robustness.
With the rise of deep learning, ~\cite{DeepBinary00} regarded the face anti-spoofing as a binary classification task and leveraged CNN to solve it. 
To avoid overfitting to the dataset, ~\cite{additionInfo01,additionInfo02,zhang2021aurora,Revisiting} utilized additional information, such as depth map, reflect map, r-ppg signal and reflect lighting, to boost the performance. 
Based on auxiliary information, some methods regularized features from the perspective of disentanglement~\cite{disentangle01,STCN}.
Some methods put forward specific convolution operators to extract spoof cues, such as CDCN~\cite{CDCN}, BCN~\cite{BCN}, DC-CDN~\cite{DCCN}, \textit{etc}. 
To achieve improvements in cross-testing, \cite{DG01,zhihong00,meta00,Domain05} adopted the strategy of domain generalization, meta-learning and few-shot learning.

However, we clarify that current methods have some shortcomings. First, the methods based on the complete face input may overfit the facial structure. 
To alleviate this issue, we put forward Structure Destruction Module and Content Combination Module to utilize the local patches. 
Second, though~\cite{additionInfo00} also utilized a two-stream CNN-based approach to extract local features and holistic depth maps, they simply fused the classification results of two branches and did not make full use of local regions. 
In contrast, we introduce Local Relation Modeling Module to tap the potential of the relationship of patches.

\begin{figure*}[t!]
    \centering
    \includegraphics[width = 0.88\linewidth]{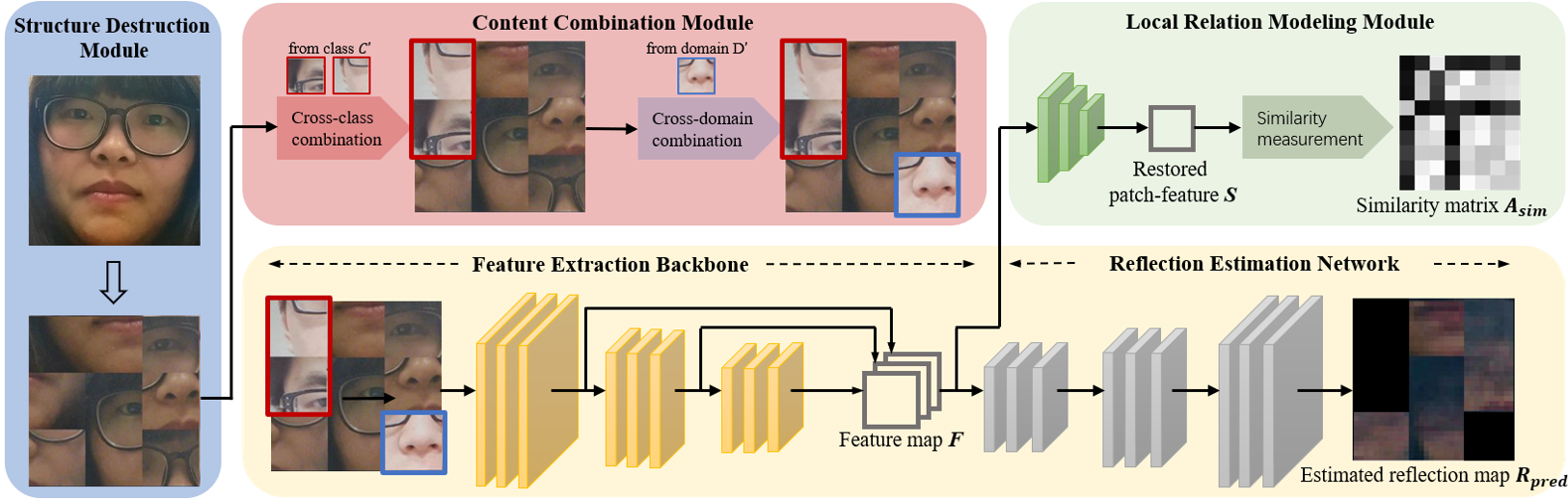}
    
    \caption{\textbf{Destruction and Combination Network (DCN)} contains Feature Extraction Backbone, Reflection Estimation Network and three novel modules, Structure Destruction Module, Content Combination Module and Local Relation Modeling Module.}
    \label{framework}
    \vspace{-4mm}
\end{figure*}

\section{Proposed Method}
In this section, we detail Destruction and Combination Network (DCN). 
As shown in Fig.~\ref{framework}, it contains a traditional Feature Extraction Backbone, Reflection Estimation Network and three modules: 1) \textbf{Structure Destruction Module} permutes the image patches as a jigsaw to destroy the face structure. 2) \textbf{Content Combination Module} splices patches from different classes and subdomains to a new image. 3) \textbf{Local Relation Modeling Module} contributes to more discriminative and general features by modeling second-order relationships between any two patches.

\subsection{Overall Framework}
In the framework, we utilize a traditional feature extraction backbone to obtain liveness feature map $F\in\mathbb{R}^{C_f\times H_f \times W_f}$, where $C_f$ is the channel number of the feature and $H_f\times W_f$ is the spatial size.
Given the feature map $F$, we use the reflection estimation network to predict the reflection map $R_{pred}$. We utilize state-of-the-art reflection estimation algorithm in~\cite{reflection} to estimate the reflection maps for present attacks. 
Then, these reflection maps serve as the ground truths of the spoof images.
As for the live images, all-zero maps are utilized instead. 
Finally we utilize the estimated reflection maps to detect spoof images. 

In order to destroy the facial structure and acquire more discriminative and general features, we propose Structure Destruction Module, Content Combination Module and Local Relation Modeling Module. The first two modules process full face images and output the permuted and spliced images which are fed into the feature extraction. The third module constrains the pair-wise relationships on the feature map $F$ for better generalization and discriminative ability. Next, we introduce the three modules separately.

\subsection{Structure Destruction Module}
As analyzed above, the network trained with original images is vulnerable to some liveness-irrelevant noise, \textit{e.g.}, facial structure information. Inspired by several methods~\cite{shen2019facebagnet, additionInfo00} that process on image patches, we destroy the facial structure to promote the generalization of the network.

Concretely, Structure Destruction Module mainly consists of two procedures, splitting and splicing. Given an aligned full RGB face image $I\in{[0,255]}^{3\times H\times W}$, we firstly divide it into an $M\times N$ grid of patches $\{I_{i,j}\}_{M\times N}$ which have the same height $h$ and width $w$, where $M=H/h$ and $N=W/w$. After acquiring $P=M\times N$ face patches, we permute all patches to form a new image $I'\in{[0,255]}^{3\times H\times W}$ as a jigsaw. Finally, these patch-permuted images are fed into the network for training.

Structure Destruction Module benefits the generalization in three ways. First, it destroys the facial structure explicitly so the network is kept away from some liveness-irrelevant noise. Additionally, it implicitly achieves the effect of the data augmentation. After dividing the original image into $P$ patches, we can obtain $P!$ different permutations. Compared with inputting the original images or face patches, taking the permutation as input brings stronger data augmentation. At last, the patch-based methods only consider the intra-patch information, while the permutation can aggregate much more inter-patch information.

\subsection{Content Combination Module}
Although Structure Destruction Module can make the network get rid of the facial structure, it does not take subdomain information into account, bringing on poor cross-domain generalization. To address these issues, we introduce the Content Combination Module which includes cross-class combination and cross-subdomain combination.\\

\vspace{-2mm}
\noindent\textbf{Cross-class combination} is to promote the discriminative ability of by splicing patches from different classes. 
As shown in the red rectangle in Fig.~\ref{framework}, for a $C$-class classification problem, given a patch-permuted image $I'$ with label $c$, we randomly pick $k\in\{k\in\mathbb{N}|0<k<P\}$ patch-permuted images from other $C-1$ classes. Then, we exchange any $k$ patches in $I'$ with the corresponding patches in the picked $k$ images. In this way, the patch-permuted image $I'$ is compounded into a multi-class image.

The cross-class combination makes mutual assistance among neighboring patches more difficult, which forces Feature Extraction Backbone to extract more discriminative features for those less discriminative patches.\\

\vspace{-2mm}
\noindent\textbf{Cross-subdomain combination} is to promote the generalization ability of the network across domains. Similar to the cross-class combination, the cross-subdomain combination compounds patches in different subdomains together. 
As shown in the blue rectangle in Fig.~\ref{framework}, for a $D$-domain generalization problem, given a patch-permuted image $I'$ in domain $d$, we randomly select $l\in\{l\in\mathbb{N}|0<l<P\}$ patch-permuted images from other $D-1$ domains. 
After that, we exchange any $l$ patches in $I'$ with the corresponding patches in the selected $l$ images. 
And then 
a patch-permuted image with multi-subdomain information is obtained.

The cross-subdomain combinations makes Feature Extraction Backbone spatially process the features from different subdomains. Compared with feeding images from different subdomains into the network one by one, the cross-subdomain combination urges the feature extractor to be domain-independent instead of covering multi-subdomains. Intuitively, the domain-independent features might perform better in unknown domains.


\subsection{Local Relation Modeling Module}
Since Content Combination Module only implicitly provides the condition of domain generalization, we still need explicit constraints to make the features in the same class more similar and the features in different classes less similar. Hence, we propose Local Relation Modeling Module which consists of a patch-feature restoration network and pair-wise similarity measurement.\\

\vspace{-2mm}
\noindent\textbf{Patch-Feature Restoration Network} attains a feature map $S\in\mathbb{R}^{C_s \times M\times N}$ from the feature map $F$, where $C_s$ is the channel number and $M\times N$ is the spatial size of the feature. 
Due to the spatial invariance of CNN, the feature vector $S_{i,j}\in\mathbb{R}^{C_s}$ on the coordinate $(i,j)$ can be regarded as the feature of the corresponding patch $I'_{i,j}$.\\

\vspace{-2mm}
\noindent\textbf{Pair-wise similarity measurement} is to constrain the similarities between every two patches. In detail, after obtaining the feature of each patch, we utilize cosine distance to measure the similarity of two patches. Given a feature $S_{i,j}$ in any coordinate of $S$ and a feature $S_{i',j'}$ in another coordinate of $S$, their cosine similarity is computed via:
\begin{equation}
    cos(S_{i,j}, S_{i',j'}) = \frac{S_{i,j}\cdot S_{i',j'}}{\left\| S_{i,j}\right\|_2 \cdot \left\| S_{i',j'}\right\|_2},
\end{equation}

\noindent In this way, we can get the cosine similarity for any two patches so that a $P\times P$ similarity matrix $A_{sim}\in[-1,1]^{P\times P}$ is obtained and $A_{sim(i,j)}=cos(S_{i/N+1, i\%N},S_{j/N+1, j\%N})$. 


For an original image or a patch-permuted image, the similarity matrix $A_{sim}$ is supposed to be an all-one matrix, because the patches are from the same class. 
While after introducing Content Combination Module, the elements of the similarity matrix become different, since patches from various classes and domains are compared together. 
Considering that the reflection map can't restrict the similarity matrix $A_{sim}$ very well, we establish an explicit relationship between any two patches and we assume that the cosine similarity of different classes should be close to -1 and the cosine similarity of the same class should be close to 1. During training, we construct a similarity matrix label $A_{label}$ for each sample, which is composed of 1 and -1 as described above. Then a pair-wise loss function is proposed:
\begin{equation}
    \vspace{-2mm}
    \mathcal{L}_{sim} = \frac{1}{P(P-1)} \left\| A_{sim} - A_{label} \right\|_{s}^2,
\end{equation}

\begin{table}[t!]
  \tiny
  \centering
  \caption{\textbf{The intra-testing results on Oulu-NPU}.}
    \linespread{1}\selectfont
    \resizebox{0.40\textwidth}{!}{
    \begin{tabular}{|c|c|c|c|c|}
    \hline
    Prot. & Method & APCER(\%) & BPCER(\%) & ACER(\%) \\
    \hline
    \multirow{7}*{1}    & FaceDe-S    & 1.2   & 1.7   & 1.5 \\ \cline{2-5}
                        & FAS-TD      & 2.5   & 0.0   & 1.3 \\ \cline{2-5}
                        & Disentangle & 1.7   & 0.8   & 1.3 \\ \cline{2-5}
                        & STDN        & 0.8   & 1.3   & 1.1 \\ \cline{2-5}
                        & CDCN        & 0.4   & 1.7   & 1.0 \\ \cline{2-5}
                        & BCN         & \textbf{0.0}   & 1.6   & 0.8 \\ \cline{2-5}
                        & Ours        & 1.3   & \textbf{0.0}   & \textbf{0.6} \\ \cline{2-5}
    \hline
    \hline
    \multirow{7}*{2}    & Disentangle & 2.7   & 2.7   & 2.4 \\ \cline{2-5}
                        & STASN       & 4.2   & 0.3   & 2.2 \\ \cline{2-5}
                        & FAS-TD      & 1.7   & 2.0   & 1.9 \\ \cline{2-5}
                        & STDN        & 2.3   & 1.6   & 1.9 \\ \cline{2-5}
                        & BCN         & 2.6   & 0.8   & 1.7 \\ \cline{2-5}
                        & CDCN        & 1.5   & 1.4   & \textbf{1.5} \\ \cline{2-5}
                        & Ours        & 2.2   & 2.2   & 2.2 \\ \cline{2-5}
    \hline
    \hline
    \multirow{7}*{3}    & Auxliary      & 2.7$\pm$1.3 & 3.1$\pm$1.7 & 2.9$\pm$1.5 \\ \cline{2-5}
                        & STDN          & 1.6$\pm$1.6 & 4.0$\pm$5.4 & 2.8$\pm$3.3 \\ \cline{2-5}
                        & STASN         & 4.7$\pm$3.9 & 0.9$\pm$1.2 & 2.8$\pm$1.6 \\ \cline{2-5}
                        & BCN           & 2.8$\pm$2.4 & 2.3$\pm$2.8 & 2.5$\pm$1.1 \\ \cline{2-5}
                        & CDCN          & 2.4$\pm$1.3 & 2.2$\pm$2.0 & 2.3$\pm$1.4 \\ \cline{2-5}
                        & Disentangle   & 2.8$\pm$2.2 & 1.7$\pm$2.6 & 2.2$\pm$2.2 \\ \cline{2-5}
                        & Ours          & 2.3$\pm$2.7 & \textbf{1.4$\pm$2.6} & \textbf{1.9$\pm$1.6} \\ \cline{2-5}
    \hline
    \hline
    \multirow{7}*{4}    & STASN        & 6.7$\pm$10.6  & 8.3$\pm$8.4    & 7.5$\pm$4.7 \\ \cline{2-5}
                        & CDCN         & 4.6$\pm$4.6   & 9.2$\pm$8.0    & 6.9$\pm$2.9 \\ \cline{2-5} 
                        & FaceDe-S     & 1.2$\pm$6.3   & 6.1$\pm$5.1    & 5.6$\pm$5.7 \\ \cline{2-5}
                        & BCN          & 2.9$\pm$4.0   & 7.5$\pm$6.9    & 5.2$\pm$3.7 \\ \cline{2-5}
                        & Disentangle  & 5.4$\pm$2.9   & 3.3$\pm$6.0    & 4.4$\pm$3.0 \\  \cline{2-5}
                        & STDN         & \textbf{2.3$\pm$3.6}   & 5.2$\pm$5.4    & 3.8$\pm$4.2 \\  \cline{2-5}
                        & Ours         & 6.7$\pm$6.8   & \textbf{0.0$\pm$0.0}    & \textbf{3.3$\pm$3.4} \\  \cline{2-5}
    \hline
    \end{tabular}
    }
  \label{Oulu-NPU}
  \vspace{-4mm}
\end{table}

\subsection{Loss Functions}
There are two loss functions in our framework, the pair-wise loss and the reflection map estimation loss. For the latter, the patches from the original reflection labels should be permuted and spliced along with the patches in the original RGB images. 
Given the permuted and spliced reflection map label $D_{label}$ and the predicted reflection map $R_{pred}$, the reflection map estimation loss is formulated as Eq.~\ref{loss_depth}.

The overall loss function Eq.~\ref{loss_sum}, is the combination of the pair-wise loss function and the reflection map estimation loss function, where $\lambda_1$ is the weight to balance two losses.

\begin{equation} \label{loss_depth}
    \setlength{\abovedisplayskip}{-2pt}
    \mathcal{L}_{reflection} = \frac{1}{H_f\times W_f} \left\| R_{pred} - R_{label} \right\|_{F}^2.
    \setlength{\belowdisplayskip}{-1pt}
\end{equation}

\begin{equation} 
\setlength{\abovedisplayskip}{-2pt}
\begin{split}
    {\mathcal{L}_{overall}} = 
    {\mathcal{L}_{sim}} + \lambda_1{\mathcal{L}_{reflection}}.
\end{split}
\setlength{\belowdisplayskip}{-1pt}
\label{loss_sum}
\end{equation}


\section{Experiments}
\subsection{Experimental Setting}
\noindent \textbf{Datasets.}
Four public databases are tested: Oulu-NPU~\cite{Oulu}, SiW~\cite{additionInfo01}, CASIA-MFSD~\cite{CASIA} and Replay-Attack~\cite{Replay}. The intra-testing performance is conducted on Oulu-NPU and SiW datasets, and the cross-testing results are reported on Replay-Attack and CASIA-MFSD datasets.

\noindent \textbf{Metrics.}
We utilize the same metrics as the previous works: Attack Presentation Classification Error Rate (\textsl{APCER})~\cite{metrics}, Bona Fide Presentation Classification Error Rate (\textsl{BPCER})~\cite{metrics}, Average Classification Error Rate (\textsl{ACER}) = (\textsl{APCER+BPCER})/2~\cite{metrics} and Half Total Error Rate (\textsl{HTER}) = (FAR + FRR)/2~\cite{metrics}.

\noindent  \textbf{Implementation Details.}
We utilize a face detector or face location files provided in datasets to extract the faces and resize them to 255 $\times$ 255. Models are implemented via Pytorch~\cite{pytorch} and trained with batch size of 20 on GeForce GTX 1080. For each mini-batch, we randomly select negative images and positive images with the ratio 1:1. For training, we set learning rate of 1e-5 with Adam Optimizer~\cite{adam} and $\lambda_1$ in Eq.~\ref{loss_sum} as 10.

\begin{table}[t!]
  \tiny
  \centering
  \caption{\textbf{The intra-testing results on SiW}.}
    \linespread{1.1}\selectfont
    \resizebox{0.40\textwidth}{!}{
    \begin{tabular}{|c|c|c|c|c|}
        \hline
        Prot. & Method & APCER(\%) & BPCER(\%) & ACER(\%) \\
        \hline
        \multirow{7}*{1}    & Auxiliary   & 3.6 & 3.6 & 3.6 \\ \cline{2-5}
                            & Meta-FAS-DR & 0.5 & 0.5 & 0.5 \\ \cline{2-5}
                            & BCN         & 0.6 & 0.2 & 0.4 \\ \cline{2-5}
                            & Disentangle & 0.1 & 0.5 & 0.3 \\ \cline{2-5}
                            & CDCN        & 0.1 & 0.2 & 0.1 \\ \cline{2-5}
                            & STDN        & 0.0 & 0.0 & 0.0 \\ \cline{2-5}
                            & Ours        & \textbf{0.0} & \textbf{0.0} & \textbf{0.0} \\ \cline{2-5}
        \hline
        \hline
        \multirow{7}*{2}    & Auxiliary   & 0.6$\pm$0.7 & 0.6$\pm$0.7 & 0.6$\pm$0.7 \\ \cline{2-5}         
                            & Meta-FAS-DR & 0.3$\pm$0.3 & 0.3$\pm$0.3 & 0.3$\pm$0.3 \\ \cline{2-5}
                            & BCN         & 0.1$\pm$0.2 & 0.2$\pm$0.0 & 0.1$\pm$0.1 \\ \cline{2-5}
                            & Disentangle & 0.1$\pm$0.2 & 0.1$\pm$0.1 & 0.1$\pm$0.0 \\ \cline{2-5}
                            & CDCN        & 0.0$\pm$0.0 & 0.1$\pm$0.1 & 0.1$\pm$0.0 \\ \cline{2-5}
                            & STDN        & 0.0$\pm$0.0 & 0.0$\pm$0.0 & 0.0$\pm$0.0 \\ \cline{2-5}
                            & Ours        & \textbf{0.0$\pm$0.0} & \textbf{0.0$\pm$0.0} & \textbf{0.0$\pm$0.0} \\ \cline{2-5}
        \hline
        \hline
        \multirow{7}*{3}    & Auxiliary   & 8.3$\pm$8.3 & 8.3$\pm$8.3 & 8.3$\pm$8.3 \\ \cline{2-5}
                            & STDN        & 8.3$\pm$3.3 & 7.5$\pm$3.3 & 7.9$\pm$3.3 \\ \cline{2-5}
                            & Meta-FAS-DR & 8.0$\pm$5.0 & 7.4$\pm$5.7 & 7.7$\pm$5.3 \\ \cline{2-5}
                            & Disentangle & 9.4$\pm$6.1 & 1.8$\pm$2.6 & 5.6$\pm$4.3 \\ \cline{2-5}
                            & BCN         & 2.6$\pm$0.9 & 2.3$\pm$0.5 & 2.5$\pm$0.7 \\ \cline{2-5}
                            & CDCN        & \textbf{1.7$\pm$0.1} & \textbf{1.8$\pm$0.1} & \textbf{1.7$\pm$0.1} \\ \cline{2-5}
                            & Ours        & 3.8$\pm$4.3 & 3.0$\pm$2.6 & 3.4$\pm$0.9 \\ \cline{2-5}
        \hline
      \end{tabular}
      }
      \label{SiW}
      \vspace{-4mm}
\end{table}

\subsection{Experimental Comparison}

\subsubsection{Intra-Testing.}
We follow the protocols defined in datasets to conduct the experiments. Tab.~\ref{Oulu-NPU} shows the comparison of our method and other methods on Oulu dataset.
Our method achieves the best performance in protocols 1, 3, 4, and gets slightly worse results in protocol 2. With a simpler backbone, our method gets better results than the methods utilizing more auxiliary information or specific structure, confirming that our method has better representative capability if all methods are of the same model capacity.
Tab.~\ref{SiW} demonstrates the results on SiW dataset. Our method performs better in protocols 1, 2 and gets comparable results in protocol 3.

\subsubsection{Cross-Testing.}
As illustrated in Tab.~\ref{CASIA-Replay}, the generalization capability is evaluated on CASIA-MFSD and Replay-Attack in terms of HTER. For a fair comparison, only the methods based on single frame information are included. 
From CASIA-MFSD to Replay-Attack, we achieve progress over the previous best model by 0.2\% on HTER. While from Replay-Attack to CASIA-MFSD, our method still approaches the prior state-of-the-art. 
The overall cross-dataset results verify that our method achieves desirable generalization capability without specific auxiliary supervision.

\begin{table}[t!]
  \tiny
  \centering
  \caption{\textbf{The cross-testing results on CASIA and Replay}.}
    \linespread{1.1}\selectfont
    \resizebox{0.40\textwidth}{!}{
    \begin{tabular}{|c|c|c|c|c|}
        \hline
        \multirow{3}*{Method}  & Train & Test   & Train  & Test   \\ \cline{2-5}
                              & CASIA & Replay & Replay & CASIA  \\
                              & MFSD  & Attack & Attack & MFSD   \\
        \hline
        Motion-Mag        & \multicolumn{2}{c|}{50.1\%} & \multicolumn{2}{c|}{47.0\%} \\ \hline
        Spectral cubes    & \multicolumn{2}{c|}{34.4\%} & \multicolumn{2}{c|}{50.0\%} \\ \hline
        LowPower          & \multicolumn{2}{c|}{30.1\%} & \multicolumn{2}{c|}{35.6\%} \\ \hline
        CNN               & \multicolumn{2}{c|}{48.5\%} & \multicolumn{2}{c|}{45.5\%} \\ \hline
        STASN             & \multicolumn{2}{c|}{31.5\%} & \multicolumn{2}{c|}{30.9\%} \\ \hline
        Auxiliary         & \multicolumn{2}{c|}{27.6\%} & \multicolumn{2}{c|}{\textbf{28.4\%}} \\ \hline
        BASN              & \multicolumn{2}{c|}{23.6\%} & \multicolumn{2}{c|}{29.9\%} \\ \hline
        Disentangle       & \multicolumn{2}{c|}{22.4\%} & \multicolumn{2}{c|}{30.3\%} \\ \hline
        BCN               & \multicolumn{2}{c|}{16.6\%} & \multicolumn{2}{c|}{36.4\%} \\ \hline
        CDCN              & \multicolumn{2}{c|}{15.5\%} & \multicolumn{2}{c|}{32.6\%} \\ \hline
        Ours              & \multicolumn{2}{c|}{\textbf{15.3\%}} & \multicolumn{2}{c|}{\underline{29.4\%}} \\ \hline
        \end{tabular}
    }
        \label{CASIA-Replay}
\end{table}

\begin{table}[t!]
  \tiny
  \centering
  \caption{\textbf{The ablation study on each module.} 
  Blank cells indicates the corresponding module is NOT deployed.}
    \resizebox{0.40\textwidth}{!}{
    \linespread{1.1}\selectfont
         \begin{tabular}{|c|c|c|c|}
         \hline
         Structure  & Content  & Local Relation  & \multirow{2}*{ACER(\%)} \\ 
         Destruction & Combination & Modeling & \\
         \hline
            &    &    &  1.67  \\
         \hline
          \checkmark  &    &    & 1.34  \\
        \hline
            & \checkmark   &    & 1.15  \\
        \hline
          \checkmark  & \checkmark   &    & 0.94  \\
        \hline
            & \checkmark   & \checkmark   & 0.73  \\
        \hline
          \checkmark & \checkmark & \checkmark & \textbf{0.62}  \\
         \hline
         \end{tabular}
         }
      \label{ablation}
      \vspace{-5mm}
\end{table}


\subsection{Ablation Study}
In order to exploit the effect of Structure Destruction Module, Content Combination Module and Local Relation Modeling Module, we first conduct ablation experiments on Oulu-NPU protocol 1, respectively. Then, we go deep into each module for verification.
The ablation results are shown in Tab.~\ref{ablation}, illustrating the contribution of each component.

\begin{figure}[t!]
    \centering
    \includegraphics[width = 0.90\linewidth]{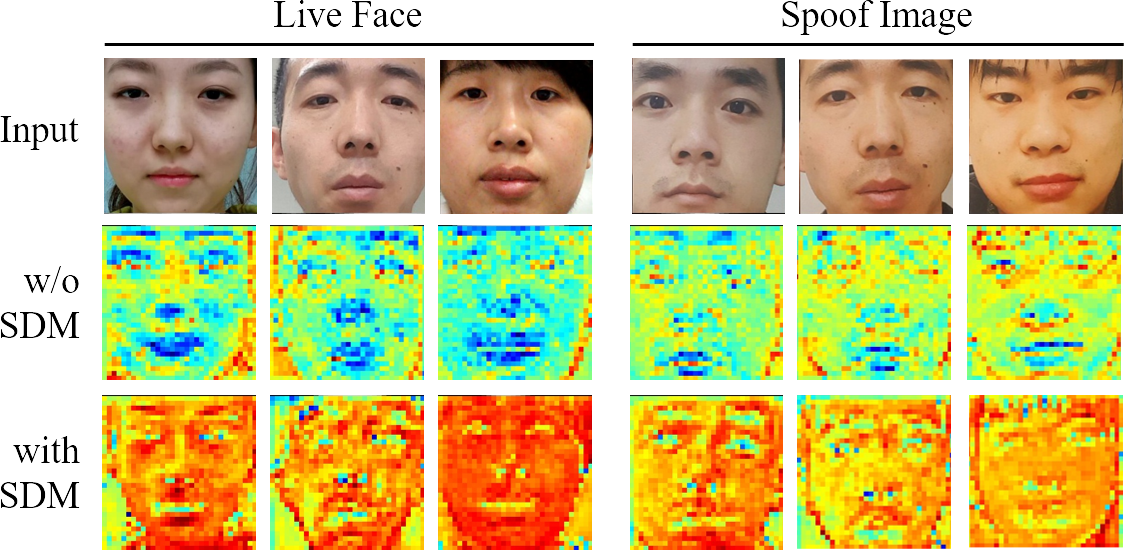}
    \caption{\textbf{The feature maps of the method with/without Structure Destruction Module.} }
    \vspace{-1mm}
    \label{featuremap}
\end{figure}

\textbf{Structure Destruction Module.}
In this section, we visualize the feature maps of the method trained with/without SDM, which are visualized in Fig.~\ref{featuremap}. The images in the second row are from the method without SDM and the third row are from the method with SDM. 
The feature maps without SDM have a high response to the structural information, such as the edges of faces. 
While the feature maps with SDM have a much more balanced response to the whole face, which verifies that SDM indeed alleviates the structural information.


\begin{figure}[t!]
    \centering
    \includegraphics[width = 0.90\linewidth]{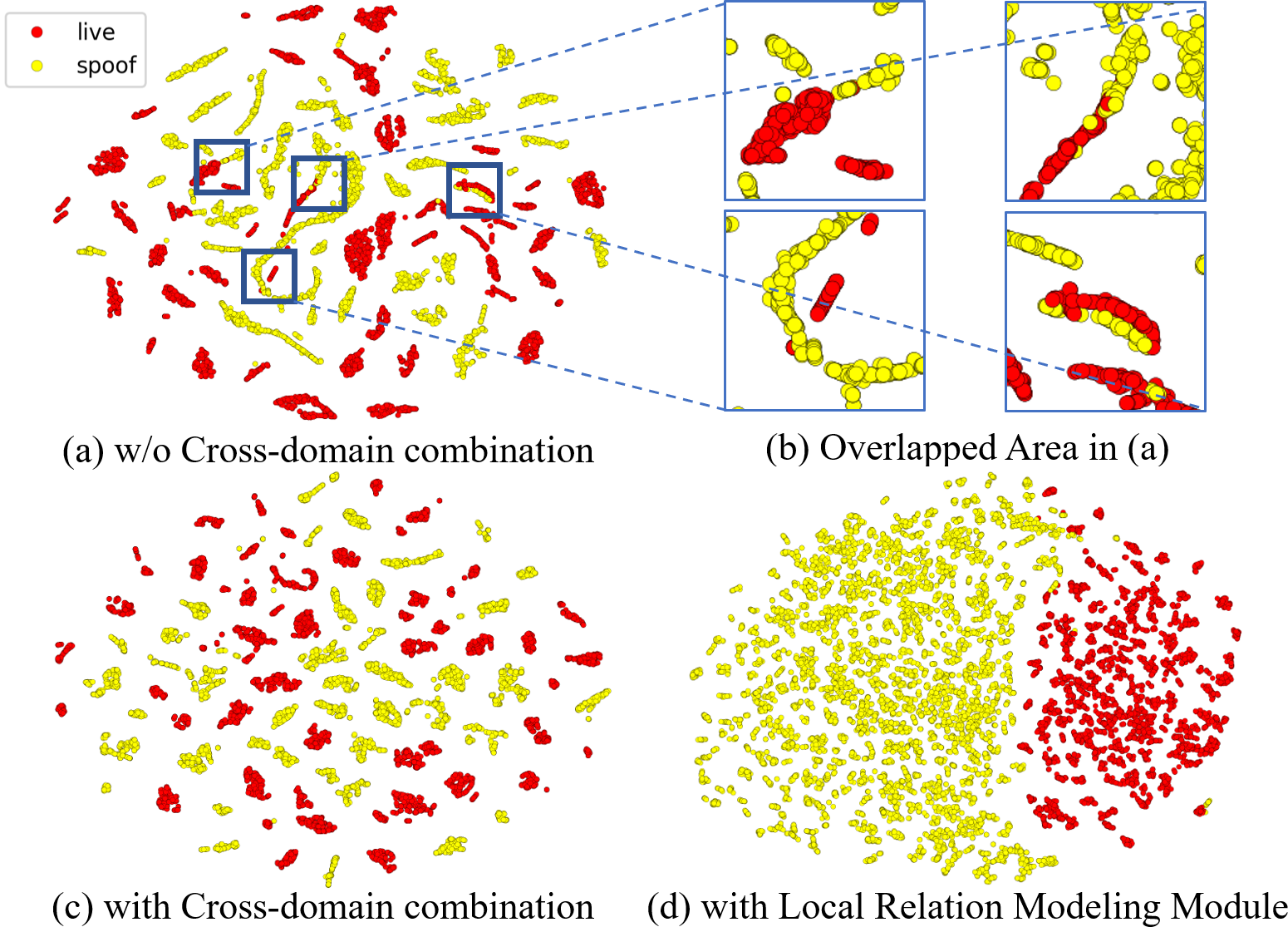}
    \caption{\textbf{The feature distributions vary with modules.} }
    \vspace{-3mm}
    \label{combination_similarity}
\end{figure}

\textbf{Content Combination Module.}
\label{Content Combination Module}
In this section, we verify the importance of two combinations in Content Combination Module (CCM) respectively. 




We utilize t-SNE~\cite{TSNE} to visualize the feature distributions from the methods trained on Replay and tested on CASIA, shown in Fig.~\ref{combination_similarity} (a), (b) and (c). The features without cross-subdomain combination are overlapped between live and spoof in the blue rectangles, while the features with such combination are separated clearly. This certifies that cross-subdomain combination promotes the generalization ability of the network across different domains.

\textbf{Local Relation Modeling Module.} 
To elucidate the effectiveness of Local Relation Modeling Module (LRMM), we display the feature distributions with/without LRMM. 
We extract the features on Oulu protocol 1, which are further visualized by t-SNE~\cite{TSNE} in Fig.~\ref{combination_similarity}(d). Compared with the features in Fig.~\ref{combination_similarity}(c), the features with LRMM are much more distinguishable. Thus, LRMM indeed aggregates the features of the same class and separates the ones of the different classes.




\section{Conclusion}
In this paper, we introduce a novel Destruction and Combination Network (DCN) to tackle the face anti-spoofing problem.
The proposed Structure Destruction Module alleviates the influence of facial structure information, and Content Combination Module introduces more variations into one image.
Then, Region Relation Learning distills the most transferable features from generated images and regularizes the feature space for Reflection Branch. 
Extensive experiments exhibit the superior performance of our method against the state-of-the-art competitors.

{\small
\bibliographystyle{ieee}
\bibliography{submission_example}
}

\end{document}